\documentclass{article}

\begin{document}

\title{Learning Probabilistic Programs Using Backpropagation}
\author{Avi Pfeffer \\ Charles River Analytics}
\maketitle

\begin{abstract}

Probabilistic modeling enables combining domain knowledge with learning from data, thereby supporting learning from fewer training instances than purely data-driven methods. However, learning probabilistic models is difficult and has not achieved the level of performance of methods such as deep neural networks on many tasks. In this paper, we attempt to address this issue by presenting a method for learning the parameters of a probabilistic program using backpropagation. Our approach opens the possibility to building deep probabilistic programming models that are trained in a similar way to neural networks.

\end{abstract}

\section{Introduction}

One of the biggest attractions of probabilistic modeling is that it enables you to combine domain knowledge with learning from data. Neural networks have proven to be extremely effective at learning but usually require a lot of data. The ability to incorporate explicit domain knowledge could significantly reduce the amount of data required. Another appealing property of probabilistic models is that they can be interpretable and are often easier to explain than neural networks. Probabilistic programming provides a modular, composable, and expressive framework for representing probabilistic models. However, probabilistic programs have generally been considered not as good as neural networks as learning machines, possibly because inference and learning of complex models is quite hard. This paper is an attempt to address this issue by providing a way to learn probabilistic programs using back-propagation, similar to neural networks. The hope is that this will provide probabilistic programs with similar scalability properties to neural networks. A second benefit is that we provide a framework that allows the entire range of possibilities, from purely knowledge-driven to purely data-driven, and everything in between, to be represented, reasoned with, and learned in a coherent way.

This work builds on that of Binder et al.~\cite{russell1995local, binder1997adaptive} (along with Buntine~\cite{buntine1994operations} and Thiesson~\cite{thiesson1997score}), who showed how to learn the conditional probability distributions of a Bayesian network with known structure and incomplete observations using gradient descent. Binder et al. also showed how to use parameterized representations of CPTs, such as noisy-or and linear Gaussians, and they also showed how to handle situations where parameters are used in more than one CPT, such as dynamic models. Their work has everything needed to learn a probabilistic program using gradient descent.

Unfortunately, Binder et al's method requires performing inference to compute the gradient. Inference can be expensive in probabilistic models and is prohibitive as the inner loop of a gradient descent algorithm. As a result, their method has largely been ignored for learning Bayesian networks, with most practitioners using EM. EM also requires inference at each step, but usually requires far fewer iterations than gradient descent to converge. However, EM is also considered a slow algorithm that is ultimately unsatisfying.

In this paper, we describe a method for learning the parameters of a probabilistic program using gradient descent in a way that does not require inference. All that is required is the ability to generate samples from the prior distribution of the program, which is easy. In each iteration of the algorithm, a number of samples is generated. Only the subset of the program necessary for generation of the samples is expanded. These samples are an approximate representation of the distribution defined by the program. A loss function that compares the empirical data distribution to this distribution is formulated. The gradient of this loss with respect to the program parameters is computed using reverse-mode automatic differentiation, or backpropagation.

We are releasing an early version of this paper in the hope of generating discussion and getting feedback.

\section{Related Work}

Generative adversarial nets (GANs)~\cite{goodfellow2014generative} are generative neural networks that have demonstrated considerable success at unsupervised learning of generative probabilistic models from rich data sets. However, GANs are not interpretable and it can be hard to include prior knowledge in the network, unlike a probabilistic program. Also, a minor point is that GANs don't support inference directly, but inference can be performed with an auxiliary network.

Stan~\cite{carpenter2016stan} is based on reverse-mode automatic differentiation, like ours. However, Stan uses this for Hamiltonian Monte Carlo, whereas we use it for back-propagation. Stan also requires a more restricted form of model than we do. Stan requires that the model be differentiable with respect to the variables in the model. In our framework, the model can be non-differentiable and even discontinuous with respect to the variables. It need only be differentiable with respect to the learnable parameters of the model. This makes it applicable to general-purpose probabilistic programming languages like Church.

Edward~\cite{tran2016edward} is a probabilistic programming language that enables explicit representation of inference models. These inference models are implemented in Tensor Flow, enabling many of the scalability benefits of neural networks. The main difference between Edward and our approach is that Edward requires the inference model to be written explicitly whereas our approach is black box, with inference being worked out automatically. Having to specify an inference model can be an advantage, because it enables you to encode algorithms (e.g. variational methods) that couldn't be easily derived. However, it might be hard to scale to more complex models and would be particularly difficult for non-experts to use.

Le, Beydin, and Wood~\cite{le2016inference} use neural networks to support inference in probabilistic programs, specifically to help create good proposals for sequential importance sampling. Their work differs from ours in that they do not learn the probabilistic program itself but rather an auxiliary network that assists in inference.

It is naturally possible to give ordinary neural networks a Bayesian interpretation, as in Bayesian deep learning~\cite{wang2016towards}.
A variety of specific forms of Bayesian generative neural network frameworks trainable by backpropagation have been developed, such as deep belief nets~\cite{hinton2006fast}, deep Boltzmann machines~\cite{salakhutdinov2009deep}, deep generative models~\cite{rezende2014stochastic}, and deep generative stochastic networks~\cite{bengio2014deep}. In contrast to our approach, each of these frameworks defines a specific kind of neural network with a given structure. In contrast, ours is a general framework that applies to programs in a generic probabilistic programming language.

\section{Overview of Approach}

We start with a functional probabilistic programming language like Church or Figaro. Some of the parameters of the language are designated to be learned by our method. For example, if we have an expression $\mbox{flip}(p)$, we might say that $p$ is a learnable parameter. Similarly, if we have an expression representing a conditional probability distribution between two categorical variables, the entries in the conditional probability table could be learnable parameters. The parameters of continuous distributions, such as the mean and variance of a Gaussian, could also be learnable.
Optionally, a prior-like function can be provided for each learnable parameter. This prior should be interpreted as a regularization function, where higher values of this prior are preferred. 

We require that for every construct in the language, three things can be done: (1) sample the value of the expression given its arguments; (2) compute the conditional density of a value given its arguments; (3) compute the derivative of the conditional density with respect to the parameter.
The conditional density needs to be differentiable with respect to the parameter but not with respect to the variables in the model. 

A probabilistic program defines a probability density function over values produced by the program. We define an error function on a program that measures, loosely speaking, how well this density function models a training set. We use stochastic gradient descent to learn the parameters of the program that minimize this error function.

In each iteration of gradient descent, we generate a set of samples from the distribution defined by the program, using the current parameter values. We estimate the error function using these samples. In addition, because we are only generating a specific set of samples, we can restrict out attention to expressions that are evaluated to generate these samples, rather than the full set of expressions that can be evaluated, which can be infinite. The result of this sampling is a data structure we call a parameterized probabilistic network (PPN).

We then use reverse mode automatic differentiation on this PPN to compute the gradient. The result is a backpropagation algorithm. It turns out that to compute this derivative exactly, we need to compute the derivative of the conditional density of a node given an ancestor set, and some analysis is needed to determine exactly which conditional densities need to be computed. As an approximation, however, we can assume that the parents of a node are independent for the purpose of derivative computation, which results in a simple and efficient algorithm.

\section{Technical Details}

\subsection{Representation}
We assume some Church-like functional probabilistic programming language in which some of the parameters are designated to be learned by our method. We require that all primitives of the language be differentiable with respect to the parameters and that we have a way to compute densities and derivatives using those primitives. The goal of learning will be to minimize some loss function with respect to these parameters. It turns out that traditional loss functions like negative log likelihood of the data with respect to the model don't work well with this method, but natural loss functions can be defined.

We will use the probabilistic program to create a network trainable using backpropagation. We call this network a parameterized probabilistic network (PPN). A PPN consist of a directed acyclic graph. 
Associated with each node $Y$ is a conditional probability density function $p(Y | \mathbf{X})$, where $\mathbf{X}$ are the parents of $Y$ in the graph. $Y$ must specify a way to sample values of $Y$ given $\mathbf{X}$, and a function $f_Y(\mathbf{x}, y; \theta_Y) = p(y | \mathbf{x})$ that is differentiable in the parameter $\theta_Y$. We are also given a way to calculate $\frac{\partial f_Y(\mathbf{x}, y; \theta_Y)}{\partial \theta_Y}$ for any value of $\mathbf{X}$ and $Y$. In addition, the parameters $\theta_Y$ also have a (possibly unnormalized) prior $p(\theta_Y)$. This prior should be thought of as a regularization function rather than a true prior in the Bayesian sense; $p(\theta_Y)$ indicates how much a particular value of $\theta_Y$ is preferred. If unspecified, $p(\theta_Y) = 1$.

Any sink of the graph $Z$ may be an output node. There may be multiple output nodes. Let $\mathbf{Z}$ represent all the output nodes.
For each output node $Z$, we are given a local error function $E_Z(z)$. The error is assumed to be (a) additive across different values $z$ of $Z$; and (b) additive across the output nodes $\mathbf{Z}$. The overall loss of the network is defined to be the expected value of this error, regularized by the priors of the parameters. Due to linearity of expectation, we can define the score of the network, with respect to a particular set of parameters $\Theta$, to be
\begin{eqnarray}
\label{eqn:loss}
L(\Theta) & = & \sum_{Z \in \mathbf{Z}} \int p(z) E_Z(z) dz - \sum_{Y} \log p(\theta_Y)
\end{eqnarray}

We can see now why a typical formulation of the loss as the negative log likelihood of the data given by the model cannot be used, because it does not decompose additively over the different values $z$ of $Z$. 
Instead, a natural formulation for the error function $E_Z(z)$ is the negative log probability of the value $z$ according to the data. For example, we can use Parzen windows to create a pdf from the training data and use that pdf to estimate the density of the point $z$. If we choose a sample of the data $\zeta_1, \ldots, \zeta_n$, we can define 
\begin{eqnarray}
\label{eqn:error}
E_Z(z) & = & - \log \{ \sum_{i=1}^n \frac{1}{n \sqrt{2 \pi \sigma^2}} e^{- \frac{(z - \zeta_i)^2}{2 \sigma^2}} \},
\end{eqnarray}
where $\sigma^2$ is some variance. The value of $\sigma^2$ could be determined using a validation set.

\subsection{Learning}
Our goal is to minimize the loss $L(\Theta)$ using stochastic gradient descent. 
The second term of Equation (\ref{eqn:loss}) is easy to differentiate.
\begin{eqnarray}
\label{eqn:derivPrior}
\frac{\partial \{- \sum_{Y} \log p(\theta_Y)\} }{\partial \theta_Y} & = & \frac{-p'(\theta_Y)}{p(\theta_Y)}
\end{eqnarray}
We call the first term of Equation (\ref{eqn:loss}) $E(\Theta)$.
We will differentiate $E(\Theta) = \sum_{Z \in \mathbf{Z}} \int p(z) E_Z(z) dz$ using a backpropagation algorithm.
Backpropagation requires a specific set of values for the nodes in the network. Since we have a generative model, we can sample values for all the nodes in the network. For probabilistic programming, the set of nodes is very large or potentially infinite, but a given set of samples will only instantiate a subset of the nodes, as long as the program terminates with probability one.

For forward propagation, we generate a set of $N$ samples from the probabilistic program. In the process of creating the program, we add a node to the PPN representing each expression that gets evaluated. If the same expression gets evaluated in multiple samples, it is represented by the same node. For any node $Y$, let $y_1,...,y_{n_Y}$ be the distinct values generated, let $\# Y$ be the number of samples in which the expression represented by $Y$ is evaluated, and let $\# y_i$ be the number of times value $y_i$ is generated. Similarly, for a set of nodes $\mathbf{Y} = Y^1,...,Y^m$, $\# \mathbf{y}_i$ is the number of samples in which the joint value $y_i^1, ..., y_i^m$ appears.
During the process of generation, we also create a topological order over the nodes according to the order in which they are evaluated.

To initialize the backward propagation process, we apply the error to each output node as follows:
\begin{eqnarray}
\frac{\partial E}{\partial \Theta} & = & \sum_{Z \in \mathbf{Z}}\frac{\partial \int p(z) E_Z(z) dz}{\partial \Theta}  \\
& = & \sum_Z \int E_Z(z) \frac{\partial p(z)}{\partial \Theta} dz + \sum_Z \int p(z) \frac{\partial{E_Z(z)}}{\partial \Theta} dz \\
& = & \sum_Z \int E_Z(z) \frac{\partial p(z)}{\partial \Theta} dz \\
& \simeq & \sum_Z \sum_{i = 1}^{n_Z} \frac{\# z_i}{\# Z} E_Z(z_i) \frac{\partial p(z_i)}{\partial{\Theta}}
\label{eqn:derivError}
\end{eqnarray}

The third step holds because $E_Z(z)$ does not depend on $\Theta$.
We have therefore reduced the problem to computing the derivative of the probability density of a particular value $z_i$ with respect to the parameters $\Theta$.
We actually generalize this slightly, providing a method to compute the derivative of the conditional probability density of any value of a variable given a specific value of ancestors of the variable.

Notation: Let $Y$ be any node. $\bar{\Theta}_Y$ denotes all the parameters of nodes that precede $Y$ in the ordering. For any set of ancestors $\mathbf{U}$ of $Y$, we now define a procedure to compute $\frac{\partial p(y_i | \mathbf{u}_j)}{\partial{\theta_Y}}$ and $\frac{\partial p(y_i | \mathbf{u}_j)}{\partial \bar{\Theta}_Y}$.

Let the parents of $Y$ be $\mathbf{X}$. First, we approximate
\begin{eqnarray}
\frac{\partial p(y_i | \mathbf{u}_j)}{\partial \theta_Y} 
& = & \frac{\partial \int p(\mathbf{x} | \mathbf{u}_j) p(y_i | \mathbf{x}, \mathbf{u}_j) d\mathbf{x}}{\partial \theta_Y} \\
& = & \frac{\partial \int p(\mathbf{x}| \mathbf{u}_j) f_Y(\mathbf{x}, y_i; \theta_Y) d\mathbf{x}}{\partial \theta_Y} \\
& = & \int p(\mathbf{x}| \mathbf{u}_j) \frac{\partial f_Y(\mathbf{x}, y_i; \theta_Y)}{\partial \theta_Y} d \mathbf{x} \\
& \simeq & \sum_{k = 1}^{n_\mathbf{X}}  \frac{\# \mathbf{x}_k, \mathbf{u}_j}{\# \mathbf{u}_j}\frac{\partial f_Y(\mathbf{x}_k, y_i; \theta_Y)}{\partial \theta_Y}
\label{eqn:derivSingle}
\end{eqnarray}

The second line holds because $\mathbf{X}$, being the parents of $Y$, render $Y$ conditionally independent of $\mathbf{U}$ given $\mathbf{X}$. The third line holds because $p(\mathbf{x}| \mathbf{u}_i)$ does not depend on $\theta_Y$. Because we know how to calculate $\frac{\partial f_Y(\mathbf{x}_k, y_i; \theta_Y)}{\partial \theta_Y}$, this part is done.

Next, we approximate
\begin{eqnarray}
\frac{\partial p(y_i | \mathbf{u}_j)}{\partial{\bar{\Theta}_Y}} & = & \frac{\partial \int p(\mathbf{x} | \mathbf{u}_j) p(y_i | \mathbf{x}, \mathbf{u}_j) d\mathbf{x}}{\partial \bar{\Theta}_Y} \\
& = & \frac{\partial \int p(\mathbf{x} | \mathbf{u}_j) f_Y(\mathbf{x}, y_i; \theta_Y) d\mathbf{x}}{\partial \bar{\Theta}_Y} \\
& = & \int f_Y(\mathbf{x}, y_i; \theta_Y) \frac{\partial p(\mathbf{x} | \mathbf{u}_j) }{\partial \bar{\Theta}_Y} d \mathbf{x} \\
& \simeq & \sum_{k = 1}^{n_\mathbf{X}} \frac{\# \mathbf{x}_k}{\# \mathbf{X}} f_Y(\mathbf{x_k}, y_i; \theta_Y) \frac{\partial p(\mathbf{x}_k | \mathbf{u}_j)}{\partial \bar{\Theta}_Y}
\label{eqn:derivMany}
\end{eqnarray}

We have again reduced the problem to compute the partial derivative of a conditional probability density. However, there may be more than one variable in $\mathbf{X}$ and these variables might not be conditionally independent given $\mathbf{U}$. We therefore introduce the notion of a sepset: A sepset for $\mathbf{X}$ given $\mathbf{U}$ is a set of nodes $\mathbf{V}$ such that all the variables in $\mathbf{X}$ are conditionally independent given $\mathbf{U} \cup \mathbf{V}$. Technically, any such sepset will work but we will generally prefer small sepsets.
Let $\mathbf{X} = X^1,...,X^m$.
We can now write
\begin{eqnarray}
p(\mathbf{x_i} | \mathbf{u}_j) & = & \int p(\mathbf{v} | \mathbf{u}_j) p(\mathbf{x_i} | \mathbf{v}, \mathbf{u_j}) d\mathbf{v} \\
& = & \int p(\mathbf{v} | \mathbf{u}_i) \prod_{k=1}^m p(x_i^k | \mathbf{v}, \mathbf{u_j}) d\mathbf{v} \\
& \simeq & \sum_{\ell = 1}^{n_\mathbf{V}} \frac{\# \mathbf{v}_\ell, \mathbf{u}_j}{\# \mathbf{u}_j} \prod_{k=1}^m p(x_i^k | \mathbf{v}_\ell, \mathbf{u}_j)
\label{eqn:probSepset} 
\end{eqnarray}

Therefore,
\begin{eqnarray}
\frac{\partial p(\mathbf{x}_i | \mathbf{u}_j)}{\partial \bar{\Theta}_Y} & \simeq &
\sum_{\ell = 1}^{n_\mathbf{V}} 
\frac{\# \mathbf{v}_\ell, \mathbf{u}_j}{\# \mathbf{u}_j} 
\frac{\partial \prod_{k=1}^m p(x_i^k | \mathbf{v}_\ell), \mathbf{u}_j)}{\partial \bar{\Theta_Y}} \\
& = & \sum_{\ell = 1}^{n_{\mathbf{V}}}
\frac{\# \mathbf{v}_\ell, \mathbf{u}_j}{\# \mathbf{u}_j}
\sum_{k=1}^m \left(\prod_{h=1,h \neq k}^m 
p(x_i^h | \mathbf{v}_\ell, \mathbf{u}_j)\right)
\frac{\partial p(x_i^k | \mathbf{v}_\ell, \mathbf{u}_j)}
{\partial \bar{\Theta}_Y} \\
& \simeq & 
\sum_{\ell = 1}^{n_{\mathbf{V}}}
\frac{\# \mathbf{v}_\ell, \mathbf{u}_j}{\# \mathbf{u}_j} 
\sum_{k=1}^m \left(\prod_{h=1,h \neq k}^m 
\frac{\# x_i^h, \mathbf{v}_\ell, \mathbf{u}_j}{\# \mathbf{v}_\ell, \mathbf{u}_j}\right)
\frac{\partial p(x_i^k | \mathbf{v}_\ell, \mathbf{u_j})}{\partial \bar{\Theta}_Y} \\
& = &
\sum_{\ell = 1}^{n_{\mathbf{V}}}
\left(\frac{\# \mathbf{v}_\ell, \mathbf{u}_j}{\# \mathbf{u}_j} 
\prod_{h=1}^m \frac{\# x_i^h, \mathbf{v}_\ell, \mathbf{u}_j}{\# \mathbf{v}_\ell, \mathbf{u}_j} \right)
\\ & & ~~~~
\sum_{k=1}^m
\frac{\# \mathbf{v}_\ell, \mathbf{u}_j}{\# x_i^k, \mathbf{v}_\ell, \mathbf{u}_j}
\frac{\partial p(x_i^k | \mathbf{v}_\ell, \mathbf{u}_j)}{\partial \bar{\Theta}_Y} 
\label{eqn:derivSepset}
\end{eqnarray}

The rewriting in the last line is to reduce the complexity of the algorithm by performing the multiplication separately from the initial summation.
We have now reduced the problem to computing the partial derivative of the conditional probability density of the value of any variable given a specific value of its ancestors, so we can recurse. Done!

\subsection{Complexity}

Note that for any set of variables $\mathbf{V}$, $N_{\mathbf{V}} \leq N$, where $N$ is the number of samples.
Also, let $M$ be the maximum number of parents of any node, let $K$ be the number of nodes, 
let $Q$ be the number of output nodes, and let $D$ be the number of training set examples used in an iteration of the algorithm.

Equation (\ref{eqn:derivPrior}) is computed $K$ times.
According to Equation (\ref{eqn:error}), the cost of computing $E_Z(z_i)$ is $O(D)$. Equation (\ref{eqn:derivError}) computes $E_Z(z_i)$ once for every output node $Z$ and every sample $z_i$, so the total cost of computing Equation (\ref{eqn:derivError}) is $O(QDN)$.
The cost of computing Equations (\ref{eqn:derivSingle}) and (\ref{eqn:derivMany}) is $O(N)$, while the cost of computing Equation (\ref{eqn:derivSepset}) is $O(MN)$.

The key question is how many times Equations (\ref{eqn:derivSingle}), (\ref{eqn:derivMany}), and (\ref{eqn:derivSepset}) are computed. This does not seem to be trivial to work out and there may be some graph theoretic property. We hypothesize that it's exponential in the size of the largest sepset, but analysis is needed. If the total number of times is $O(L)$, then the overall cost of performing one step of gradient descent is $O(K + QDN + LMN)$.

\subsection{Approximation}

If the complexity of computing the gradient exactly is too high, we can use an approximation scheme where we use smaller sepsets. An extreme version of this will use empty sepsets, assuming all parents are independent for the purpose of computing the gradient.
We then only need to compute the derivatives of unconditional pdfs.
Proceeding from Equation (\ref{eqn:derivError}), we compute 
\begin{eqnarray}
\frac{\partial p(y_i)}{\partial \theta_Y} & = & \frac{\int p(\mathbf{x}) p(y_i | \mathbf{x}) d\mathbf{x}}{\partial \theta_Y} \\
& \simeq & \sum_{j=1}^{n_\mathbf{X}} \frac{\# \mathbf{x}_j}{\# \mathbf{X}} \frac{\partial f_Y(\mathbf{x}_j, y_i; \theta_Y)}{\partial \theta_Y}
\label{eqn:approxDerivSingle}
\end{eqnarray}

We also compute
\begin{eqnarray}
\frac{\partial p(y_i)}{\partial \bar{\Theta}_Y} & = & \frac{\int p(\mathbf{x}) p(y_i | \mathbf{x}) d\mathbf{x}}{\partial \bar{\Theta}_Y} \\
& = &  \int f_Y(\mathbf{x}, y_i; \theta_Y) \frac{\partial p(\mathbf{x})}{\partial \bar{\Theta}_Y} d \mathbf{x} \\
& \simeq & \int f_Y(\mathbf{x}, y_i; \theta_Y) \frac{\partial (\prod_{k=1}^{m} p(x^k))}{\partial \bar{\Theta}_Y} d \mathbf{x} \\
& = & \int f_Y(\mathbf{x}, y_i; \theta_Y) \sum_{k=1}^{m} \left( \prod_{h=1, h \neq k}^m p(x^h) \right) \frac{\partial p(x^k)}{\partial \bar{\Theta}_Y} \\
& \simeq & \sum_{j=1}^{n_\mathbf{X}} f_Y(\mathbf{x}_j, y_i; \theta_Y) \sum_{k=1}^{m} 
\left(\prod_{h=1, h \neq k}^m \frac{\# x_j^h}{\# X^h}\right)  \frac{\partial p(x_j^k)}{\partial \bar{\Theta}_Y} \\
& = & \sum_{j=1}^{n_\mathbf{X}}
\left(f_Y(\mathbf{x}_j, y_i; \theta_Y) \prod_{h=1}^m \frac{\# x_j^h}{\# X^h}\right)
\sum_{k=1}^m \frac{\# X^k}{\# x_j^k} \frac{\partial p(x^k)}{\partial \bar{\Theta}_Y}
\label{eqn:approxDerivMany}
\end{eqnarray}

Equations (\ref{eqn:approxDerivSingle}) and (\ref{eqn:approxDerivMany}) are computed $O(K)$ times at a cost of $O(MN)$ each.
Therefore, the total cost of the algorithm, per iteration of gradient descent, is
$O(QDN + KMN)$.
Since the number of output nodes $Q$ is at most the total number of nodes  $K$, this can be summarized as $O(KN(D + M))$.
Thus, the complexity is linear in the number of nodes in the network, the number of samples taken, the number of training instances used per iteration, and the maximum number of parents of any node.

We note that although this scheme assumes that the parents are independent for the purposes of backpropagation, all dependencies between the parents are taken into account when generating the samples. How this approximation scheme will perform on practical problems is an open question.

Similar to frameworks for approximate inference in graphical models like mini-buckets~\cite{dechter2003mini}, we could also consider a whole range of approximations where we limit the size of sepsets that are allowed. We do not discuss this possibility further here.

\section{Discussion}

The most important questions with the method are whether it can provide similar scalability of learning to deep neural networks. Two particular issues, that have been solved for deep nets, are vanishing of the gradient as the network becomes deep and convergence of the learning to local minima. Our hope is that our use of sampled training set batches will help with the local minima problem. For the vanishing gradient problem, the key question is the sensitivity of the output of the network to parameters early in the network. If an early variable is critical to the output, the sensitivity should be high and our method should discover this. Also, as opposed to traditional neural networks, the variables closest to the data are at the end of the network, so the vanishing gradient problem may be less significant anyway.

Even after the model has been learned, using it requires inference. Inference in probabilistic models is hard, which is one advantage of neural networks. However, significant strides have been made in this area in the last few years, including using neural networks in the inference process itself. Our method is orthogonal to the work on inference; we will benefit from any advances in the inference area. One might ask, if we are relying on inference anyway, why not use full Bayesian inference and learn the parameters of the model through inference? In our experience, simultaneously learning parameters in a high dimensional space and perform probabilistic inference on the underlying model is often intractable. If the parameters can be learned separately, the underlying inference will often be much easier.

Our goal will be to evaluate our approach across a number of dimensions. First, a natural point of comparison is Stan. Stan also uses reverse mode automatic differentiation but for the purpose of inference using Hamiltonian Monte Carlo. Also, Stan implements a more limited language because model primitives need to be differentiable in the model variables, not just the parameters. We would like to compare the efficiency and accuracy of learning using our approach and Stan's and also whether our approach can scale to models that cannot easily be represented in Stan. Stan has been heavily engineered for efficiency, so any comparison will require some effort on our part to create an efficient implementation.

A second natural comparison is GANs. GANs are generative neural network models that have proven highly effective at generating data. Our hope is that by enabling domain knowledge to be explicitly included in models, it will be possible to train our models using far less data.

Ultimately, it will be interesting to see whether it is possible to represent and learn new kinds of deep probabilistic models in this framework. For example, imagine a functional LDA-style model. A topic consists of a probability distribution over transformations over some data structure, such as an image or a graph. The generative model consists of choosing a topic, then choosing a sequence of transformations from the topic, and applying them sequentially to a starting value using function composition. The transformations are all probabilistic functions that can be parameterized by parameters that are learned using the methods in this paper. This is a very general model in which LDA is a special case.

\bibliographystyle{alpha}
\bibliography{PPBackProp}

\end{document}